\documentclass[conference]{IEEEtran}
\IEEEoverridecommandlockouts
\usepackage{cite}
\usepackage{array}
\usepackage{amsmath,amssymb,amsfonts}
\usepackage{algorithmic}
\usepackage{graphicx}
\usepackage{textcomp}
\usepackage{xcolor}

\def\BibTeX{{\rm B\kern-.05em{\sc i\kern-.025em b}\kern-.08em
    T\kern-.1667em\lower.7ex\hbox{E}\kern-.125emX}}
\begin{document}

\title{A Privacy-Aware Deep Learning Benchmark for Facial Emotion Recognition \\
}

 \author{\IEEEauthorblockN{
 Larissa Palhares\IEEEauthorrefmark{1}\IEEEauthorrefmark{2}, 
 Maria Machado\IEEEauthorrefmark{1}\IEEEauthorrefmark{2}, 
 Hudson Bruno\IEEEauthorrefmark{1},
 Raphael Ramos\IEEEauthorrefmark{1},
 Andrew Vale\IEEEauthorrefmark{1},
 Barbara Barbosa\IEEEauthorrefmark{1}, \\
 Inara Souza\IEEEauthorrefmark{1}, and 
 Fabio Grassiotto\IEEEauthorrefmark{1}
 }
 \IEEEauthorblockA{\IEEEauthorrefmark{1}Eldorado Research Institute, Campinas, SP, Brazil.\\\{larissa.palhares, maria.machado, fabio.grassiotto\}@eldorado.org.br\\
 \IEEEauthorrefmark{2}Both authors contributed equally to this work.
 }
 }

\maketitle

\begin{abstract}
Facial Expression Recognition (FER) traditionally relies on raw RGB images, introducing privacy risks, sensitivity to illumination variations, and susceptibility to identity-confounding visual shortcuts. As an alternative, this work proposes a lightweight, privacy-aware pipeline using exclusively geometric representations---facial landmarks (478 points) and blendshapes (52 coefficients, ARKit standard)---eliminating persistent raw video storage. Six neural architectures trained from scratch on fused CREMA-D and RAVDESS datasets were benchmarked against an ImageNet-pretrained ResNet50 baseline under a 5-fold subject-independent cross-validation protocol. Geometric models consistently outperformed the pixel baseline on unseen subjects with 12--124 times fewer parameters, while demonstrating superior robustness on subtle emotional expressions such as sadness. Ablation studies reveal that blendshapes capture the primary discriminative signal, while landmarks provide essential complementary spatial topology. These results demonstrate that compact geometric representations preserve affective semantics, eliminate downstream transfer learning dependency on visual backbones, and enable efficient, privacy-aware emotion recognition.
\end{abstract}


\section{Introduction}
\label{sec:introduction}
In recent years, emotion recognition powered by Artificial Intelligence (AI) has established itself as a critical research domain in digital therapeutics, assistive technologies, and healthcare monitoring~\cite{arabian2024emotion, gerbaudo2026dementia}. This growing interest stems from the understanding that emotions and health are intrinsically linked in a bidirectional relationship: emotional states trigger immediate biological and physiological responses that can compromise homeostasis and contribute to long-term physical or psychological disorders~\cite{lima2025emotional}. Consequently, automated affective computing systems~\cite{picard1997affective} have made significant strides in detecting emotional patterns to aid in clinical assessment and early interventions, driven by advances in deep learning architectures~\cite{wang2022systematic}.

However, conventional Facial Expression Recognition (FER) relies predominantly on processing raw RGB images~\cite{li2022deep}, introducing severe data exposure risks under health privacy regulations (e.g., GDPR)~\cite{Price2019PrivacyIT}. Furthermore, pixel-based models often suffer from \textit{shortcut learning}~\cite{geirhos2020shortcut}: they tend to conflate affective dynamics with non-expressive confounding factors, such as subject identity, skin tone, and fluctuating illumination conditions~\cite{li2022deep}. Consequently, conventional visual backbones frequently experience significant performance degradation when evaluated in realistic, subject-independent scenarios on unseen individuals.

\begin{figure*}
\centering
\includegraphics[width=0.9\linewidth]{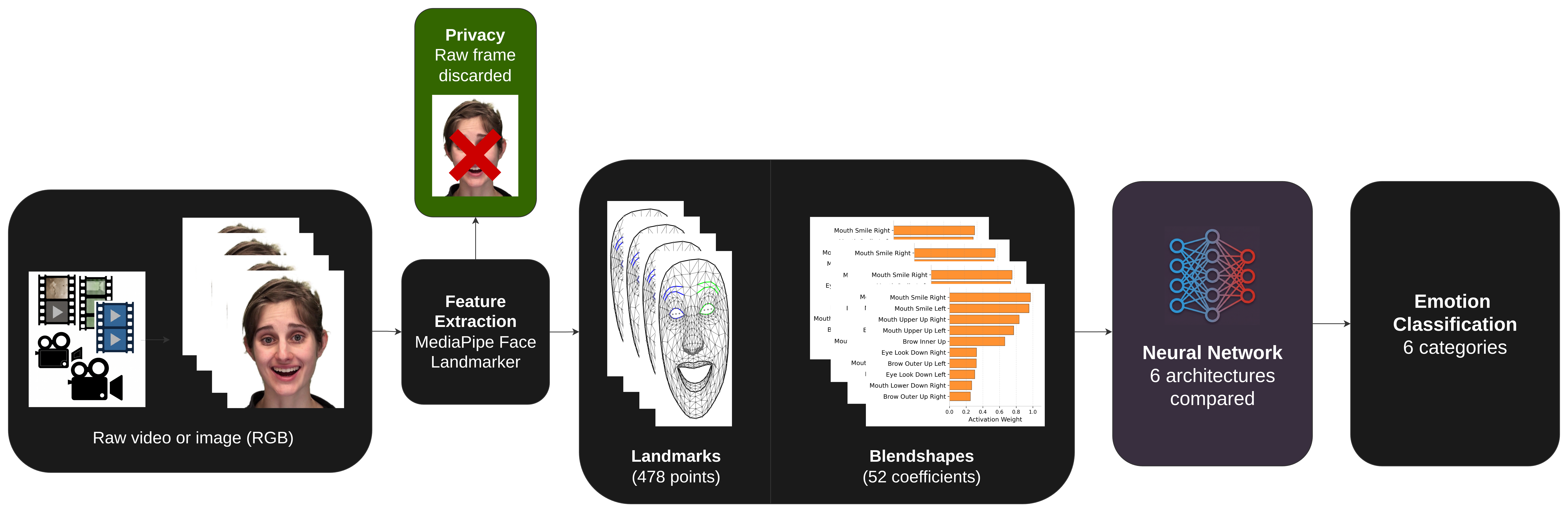}
\caption{Overview of the proposed emotion classification pipeline: facemesh landmarks ($478$ points) and blendshape coefficients ($52$ weights) are extracted via MediaPipe Face Landmarker while raw RGB frames are discarded and visually abstracted into geometric primitives to prevent biometric exposure.}
\label{fig:pipeline}
\end{figure*}

To address these limitations, abstracting facial dynamics into compact geometric representations---specifically facial landmarks~\cite{elsheikh2025improving} and blendshape coefficients~\cite{attrah2026emotion}---presents an effective paradigm shift, as illustrated in Fig.~\ref{fig:pipeline}. By avoiding raw pixel processing, these low-dimensional representations abstract away from appearance variations (e.g., skin texture and ambient lighting) while adhering to \textit{data minimization} principles through immediate frame discarding. Furthermore, delegating low-level perception to a frozen geometric front-end allows the downstream sequence classifiers to be trained entirely from scratch without task-specific visual pretraining. Despite these advantages, systematic empirical benchmarks comparing diverse deep learning families on geometric representations remain scarce.

To bridge this gap, this work presents a comprehensive benchmark of six distinct neural architectures trained from scratch on combined 3D landmark and blendshape representations, evaluating their performance and parameter efficiency against an ImageNet-pretrained ResNet50 baseline under a 5-fold subject-independent cross-validation scheme.

The main contributions of this study are:
\begin{itemize}
    \item A privacy-aware, lightweight FER pipeline operating on 3D landmarks and ARKit blendshape coefficients, eliminating the requirement for persistent raw video storage.
    \item A systematic benchmark across six distinct neural architectural paradigms (MLP, LSTM, TCN, ResNet-1D, GNN, and ST-GCN) trained from scratch on combined CREMA-D and RAVDESS datasets under subject-independent cross-validation.
    \item An empirical analysis demonstrating that compact geometric representations achieve superior generalization on unseen subjects compared to a heavy pixel-based baseline, with orders of magnitude fewer parameters.
    \item Ablation studies characterizing the complementary roles of topological landmarks and semantic blendshapes in affective classification.
\end{itemize}

The remainder of this paper is structured as follows. Section~\ref{sec:relatedwork} reviews related work on geometric emotion recognition. Section~\ref{sec:methodology} details the preprocessing pipeline and the evaluated architectures. Section~\ref{sec:results} presents the experimental benchmark, while Section~\ref{sec:ablationstudies} details the ablation studies. Finally, Section~\ref{sec:conclusion} concludes the paper and outlines future work.

\section{Related Work}
\label{sec:relatedwork}

Traditional approaches to FER rely predominantly on processing raw RGB images. However, as reviewed by Li and Deng~\cite{li2022deep}, raw pixel processing introduces vulnerability to pose variations, illumination fluctuations, and acquisition noise. Moreover, the capture and storage of raw facial footage raise critical biometric privacy concerns~\cite{Price2019PrivacyIT}. These methods also encode demographic biases inherent in training datasets~\cite{xu2020investigating, kim2021agebias}. While dataset balancing and augmentation offer limited mitigation, Hosseini et~al.~\cite{Hosseini2025FacesOF} demonstrated that residual-based Convolutional Neural Networks (CNNs) and Vision Transformers still exhibit significant demographic disparities, confirming that model-level learning remains tightly coupled to data-level biases.

In response, facial landmark extraction operates as an appearance de-identification mechanism by abstracting faces into geometric coordinates. Ju{\'a}rez-Jim{\'e}nez et~al.~\cite{juarezjimenez2025facial} evaluated normalization techniques and landmark coordinate representations, while P{\'e}rez-G{\'o}mez  et~al.~\cite{perezgomez2020feature} extracted 2D and 3D distances and angles grounded in the Facial Action Coding System (FACS) and the MPEG-4 standard, employing Principal Component Analysis (PCA) and Genetic Algorithms (GA) for feature selection, enabling ultra-lightweight models.

Complementary approaches employ blendshapes and Action Units (AUs) to capture facial muscle dynamics. Wang et~al.~\cite{wang2020leveraging} introduced a two-stream network (BlendshapeFERNet), demonstrating that blendshape deformations remain invariant to individual morphology, while Li et~al.~\cite{li2025aublendshape} showed that AU-blendshape representations are inherently less sensitive to demographic identity. Unlike conventional pixel-based architectures that are prone to encoding identity and demographic biases~\cite{Hosseini2025FacesOF}, these abstract representations effectively decouple facial muscular dynamics from static visual appearance, mitigating the learning of spurious shortcuts.

Furthermore, landmarks naturally map to graph structures. Arabian et~al.~\cite{arabian2024emotion} constructed facial meshes for Graph Convolutional Networks, while Ngoc et~al.~\cite{ngoc2020facial} proposed Directed Graph Neural Networks (GNNs) with Delaunay triangulation and temporal convolutions. Zhang et~al.~\cite{zhang2023geometric} extended these methods with distance-aware graph operations and spectral encodings for high-frequency affective dynamics.

Despite the promise of geometric FER, systematic benchmarks across diverse neural architectures remain scarce. Unlike resource-heavy Vision Transformers, lightweight sequence and topological models provide well-suited inductive biases for short temporal windows on resource-constrained devices. This work bridges this gap by benchmarking six convolutional, recurrent, and graph-based architectures trained from scratch on unified landmark and blendshape inputs, contrasting their accuracy-efficiency trade-offs against a traditional pixel baseline.

\section{Methodology}
\label{sec:methodology}

\subsection{Data Collection and Preprocessing}
\label{subsec:data_collection}

To build the dataset used in the experiments, two public datasets from the emotion recognition literature were merged: CREMA-D~\cite{cao2014cremad} and RAVDESS~\cite{livingstone2018ravdess}, both composed of videos of actors portraying controlled emotional expressions. To compose the dataset, the original videos were sampled at an extraction rate of 2 frames per second, resulting in a final combination of over 74k samples, standardized into six emotion classes: angry (18.68\%), disgust (14.85\%), fear (17.93\%), happy (17.56\%), neutral (12.14\%), and sad (18.84\%). To evaluate the models' performance and maximize data utilization, the entire dataset was subjected to a 5-fold cross-validation approach. Crucially, to prevent data leakage and ensure true generalization, these folds were created in a subject-independent manner, ensuring that actors present in the training folds were excluded from the validation fold of the same split. Consequently, the metrics reported throughout this study reflect the average performance across all five folds.

The extraction of geometric representations---facial landmarks (478 points) and blendshapes (52 coefficients, in the ARKit standard)---was performed using MediaPipe Face Landmarker~\cite{MediaPipe}, which encapsulates and accelerates the extraction pipeline, enabling efficient processing of the videos at scale.


To achieve translation and scale invariance during preprocessing, each landmark $p_i$ was spatially normalized by centering it on the nose tip ($p_{\text{nose}}$) and dividing by the Euclidean distance between the forehead ($p_{\text{forehead}}$) and chin ($p_{\text{chin}}$):

\begin{equation}
\label{eqnormaliz}
    \hat{p}_i = \frac{p_i - p_{\text{nose}}}{\Vert{}p_{\text{forehead}} - p_{\text{chin}}\Vert{}_2}.
\end{equation}

To model the temporal dynamics of facial expressions, the data were organized into fixed-length sliding windows of 5 frames, with a stride of 1. This formatting was applied uniformly across all evaluated architectures and folds, ensuring comparable input conditions.


\subsection{Evaluated Architectures}



To investigate the best way to model the spatial and temporal dependencies of facial representations (landmarks and blendshapes), six distinct geometric neural architectures---selected following a systematic literature review---were implemented. These models follow a logical progression across modeling paradigms, ranging from structure-agnostic temporal aggregation (MLP), to standard sequence modeling approaches (LSTM, TCN, ResNet-1D), and finally to architectures that explicitly encode the face's topological structure (GNN, ST-GCN).

As a traditional raw-image baseline for comparison, we evaluated a ResNet50 architecture. To reflect standard practice in the pixel-based modeling literature, this backbone was initialized with generic ImageNet weights~\cite{deng2009imagenet}. While freezing the initial layers of an ImageNet-pretrained model limits its ability to adapt to fine-grained facial morphology, this setup isolates the effect of heavy parameterization (the 25.9M parameters reported in Table~\ref{tab:results} represent the total model size) against our lightweight geometric models. The output of the spatial feature extraction for each frame feeds an LSTM layer (256 units).

The six geometric architectures evaluated are:

\begin{itemize}
    \item \textbf{Temporal MLP:} An architecture that processes landmarks and blendshapes independently, frame by frame, through a five-layer dense network (MLP). Landmarks are projected into a 512-dimensional space, while blendshapes are mapped to 128 dimensions. The features are fused into a 256-dimensional vector per frame. The temporal dependency of the sequence is then aggregated using a Masked Temporal Average Pooling strategy, ensuring that paddings are ignored, before feeding a two-layer linear classifier.
    \item \textbf{LSTM:} An architecture focused on the explicit modeling of short- and long-term sequential memory. Using the same spatial projection backbone as MLP (landmarks and blendshapes concatenated and fused into a single 256-dimensional vector per frame), the sequence feeds a single recurrent Long Short-Term Memory (LSTM) layer with a hidden size of 256 units. The video-level prediction is obtained by dynamically extracting, through length masking, the vector corresponding to the last valid frame of the sequence.

    \item \textbf{Temporal TCN:} A model grounded in Temporal Convolutional Networks~\cite{bai2018empirical}. Landmarks and blendshapes are projected separately per frame (512 and 128 dimensions, respectively) and then fused into a single 256-dimensional vector. The temporal module consists of three residual blocks with 256 channels each, using kernels of size 3 and standard dilation rates of 1, 2, and 4. While this dilated structure yields an effective receptive field larger than our fixed 5-frame sequence---meaning the convolutions operate partially over zero-padded temporal boundaries---it was maintained to preserve the standard structural capacity of the TCN baseline. The state of the last valid frame is extracted for the final classification.

    \item \textbf{ResNet-1D:} A fully convolutional approach adapted from residual architectures~\cite{hong2020holmes}. In each frame, the coordinate matrices (478 points $\times$ 3 = 1,434 values) and the 52 facial coefficients are concatenated and flattened into a unified input vector of 1,486 features (1,434 + 52). The input tensor $[B, C_{in}=1486, T=5]$ undergoes an initial $1 \times 1$ temporal convolution projecting it to 64 channels. This is followed by a hierarchy of three 1D residual blocks outputting 128, 256, and 256 channels, respectively. Each residual block contains two temporal convolutional layers with a kernel size of 3. The last block applies a stride of 2 to perform smooth temporal downsampling. Features are aggregated via Adaptive Global Average Pooling and fed into a dense classification head ($256 \rightarrow 128 \rightarrow \text{classes}$).

    \item \textbf{Temporal GNN (GIN + GRU):} A hybrid architecture that treats the anatomical coordinates of the face as an undirected spatial graph. The edge adjacency matrix is constructed using MediaPipe's canonical facial mesh tessellation. The model uses two Graph Isomorphism Network (GINConv) layers, with 64 dimensions each. Each GINConv layer employs an internal MLP ($3 \rightarrow 64 \rightarrow 64$ for the first, $64 \rightarrow 64 \rightarrow 64$ for the second) with batch normalization and ReLU activations. The 478 node embeddings are reduced to a 64-dimensional frame-level vector via global mean pooling. Concurrently, the 52 blendshape coefficients are processed by a parallel dense branch (MLP $52 \rightarrow 64 \rightarrow 32$, with ReLU). The graph and blendshape embeddings are then concatenated into a 96-dimensional per-frame vector ($64 + 32$), which feeds into a 1-layer Bidirectional GRU (128 units per direction) to model temporal evolution.

    \item \textbf{ST-GCN (Spatio-Temporal GCN):} Inspired by Spatio-Temporal Graph Convolutional Networks~\cite{yan2018spatial}, this architecture enables joint learning of anatomical and temporal dynamics. To infuse the global muscular deformations into the localized nodes, the 52-dimensional blendshape vector is linearly projected to match the internal node feature dimension (32 channels), broadcasted across all 478 nodes, and added to the projected 3D landmark coordinates. The model uses three ST-GCN blocks (32, 64, and 128 channels). Each block applies a spatial convolution (using MediaPipe's canonical topology with learnable edge importance) simultaneously with a 1D temporal convolution (kernel size of 5) along the motion trajectory of each facial point.
\end{itemize}



\subsection{Training}

A central pillar of this investigation is validating that the geometric abstraction provided by landmarks and blendshapes eliminates the need for Transfer Learning, a common practice in RGB image-based emotion recognition~\cite{bulat2022pretraining, nguyen2018meta}, where pixel-based models rely on backbones pre-trained on large image datasets (e.g., ImageNet, VGGFace) to learn low-level visual representations---edges, textures, lighting gradients, and chromatic variations---before any semantic information can be extracted.

Using landmarks and blendshapes removes low-level visual learning, as facial geometry and muscular dynamics arrive pre-extracted and semantically structured. As a direct consequence, all six architectures evaluated in this study could be randomly initialized and trained entirely from scratch, without any dependence on pre-trained weights or massive external datasets. In contrast, the pixel-based convolutional baseline (ResNet50) is a heavily parameterized deep architecture that demands substantial computational capacity. Training such a massive network from scratch on a relatively restricted dataset would not only make convergence exceedingly difficult and computationally expensive, but also inevitably lead to rapid overfitting. Therefore, it was necessary to initialize it with ImageNet pre-trained weights and fine-tune only its final layers.


To prevent overfitting and improve the generalization capability of the geometric models, a data augmentation strategy was applied exclusively to the 3D landmark coordinates, acting as a spatial regularization mechanism. This augmentation included horizontal flipping (negating the X-axis without reordering symmetric indices), in-plane geometric rotations (up to $\pm 60^\circ$), scale variations ($\pm 30\%$), and small translations. These transformations, alongside additive Gaussian noise, were applied after the initial spatial normalization (Eq.~\eqref{eqnormaliz}) to deliberately reintroduce localized spatial noise and scale variance, simulating sensor and capture imperfections. The blendshape coefficients, conversely, were left unaugmented to serve as an invariant semantic anchor. In contrast, the ImageNet-pretrained ResNet50 baseline was fine-tuned on the standard unaugmented RGB frames to isolate the effect of its pre-learned visual representations.

All six geometric architectures, as well as the ResNet50 baseline, were trained and evaluated under identical conditions using a nested subject-independent cross-validation strategy to ensure a fair comparison and prevent optimistic selection bias. For each fold, the dataset was rigorously partitioned to prevent any actor leakage across the training, internal validation, and external test sets. Specifically, 15\% of the development actors were assigned to the internal validation set. To eliminate optimizer choice as a confounding variable, the standard Adam optimizer was used uniformly across all evaluated architectures, with an initial learning rate of $1 \times 10^{-3}$ and weight decay of $1 \times 10^{-4}$. The learning rate was dynamically adjusted using a StepLR scheduler, with a decay of 0.5 every 20 epochs. The loss function used was standard Cross-Entropy. All models were trained for 50 epochs, with a fixed batch size of 32 samples. Checkpoint selection was performed exclusively on the internal validation set (best model checkpointing), and the final selected model for each fold was evaluated exactly once on the unseen external test set.


\section{Results}
\label{sec:results}

The training characteristics are directly reflected in the structural complexity of the proposed models, whose number of trainable parameters ranged from 208k to 2.15M (Table~\ref{tab:results}). This value is well below the conventional convolutional backbones used in pixel-based approaches, which typically contain tens of millions of parameters. One example is ResNet50, the raw-image baseline model tested among the architectures, which is commonly employed as a feature extractor in emotion recognition tasks via transfer learning. This architecture's implementation in this work totaled 25,870,918 parameters---more than ten times the value observed in the landmark- and blendshape-based models.

Table~\ref{tab:results} presents the average 5-fold cross-validation test accuracy and F1-Score (macro) for each architecture, along with their respective standard deviations and the number of trainable parameters. 

\begin{table}[htbp]
\caption{Subject-independent 5-fold cross-validation test performance and trainable parameters per architecture. Bold values indicate the best results.}
\label{tab:results}
\begin{center}
\resizebox{\columnwidth}{!}{%
\begin{tabular}{|l|r|c|c|} 
\hline
\textbf{Architecture} & \textbf{Parameters} & \textbf{Accuracy (\%)} & \textbf{Macro F1-Score (\%)} \\
\hline
ResNet-1D & 1,001,158 & $\mathbf{58.06 \pm 6.59}$ & $\mathbf{57.65 \pm 6.38}$ \\
Temporal GNN & \textbf{208,870} & $57.15 \pm {4.06}$ & $56.50 \pm 4.10$ \\
ST-GCN & 834,898 & $56.64 \pm 5.45$ & $56.35 \pm 5.22$ \\
Temporal TCN & 2,155,142 & $56.64 \pm 5.69$ & $55.88 \pm 6.10$ \\
LSTM & 1,500,294 & $55.86 \pm 5.88$ & $55.28 \pm 5.92$ \\
Temporal MLP & 973,958 & $55.05 \pm 6.39$ & $54.68 \pm 6.24$ \\
ResNet50 (RGB) & 25,870,918 & $51.97 \pm 6.00$ & $50.86 \pm 6.18$ \\
\hline
\end{tabular}%
}
\end{center}
\end{table}

The data show that ResNet-1D achieved the best cross-validation test accuracy (58.06\%) and the best macro F1-Score (57.65\%). GNN achieved the second-best performance (57.15\% accuracy and macro F1-Score of 56.50\%) with a much smaller fraction of trainable parameters: roughly 208k, a volume approximately 4 to 10 times smaller than that of the other landmark-based architectures. This result suggests that the explicit exploration of facial topology via GINConv may offer a more compact spatial representation compared to approaches that process landmarks in a flattened form, maintaining competitive performance with a reduced parameter count.

Among the other sequence modeling approaches, ST-GCN and TCN tied in accuracy (56.64\%), although TCN required the largest number of parameters among the landmark architectures (2.1M), showing that greater structural capacity did not necessarily translate into better generalization for this problem. MLP showed the lowest performance within the landmark-based set (55.05\% accuracy). Most notably, considering the raw-image baseline, ResNet50 (RGB) presented the lowest overall performance in the study, with 51.97\% test accuracy and a macro F1-Score of 50.86\%. This observation indicates that a larger parameter count, comprising 25.9M parameters (which represent between 12 and 124 times the volume used by the other architectures), did not guarantee better generalization under the cross-validation setup, even when utilizing representations pre-learned from external images (ImageNet). The proposed landmark- and blendshape-based architectures, on the other hand, reached higher accuracy levels (up to 58.06\%) while being trained entirely from scratch and inherently preserving privacy at the data source.

\begin{figure*}[htbp]
\centering
\includegraphics[width=0.8\linewidth]{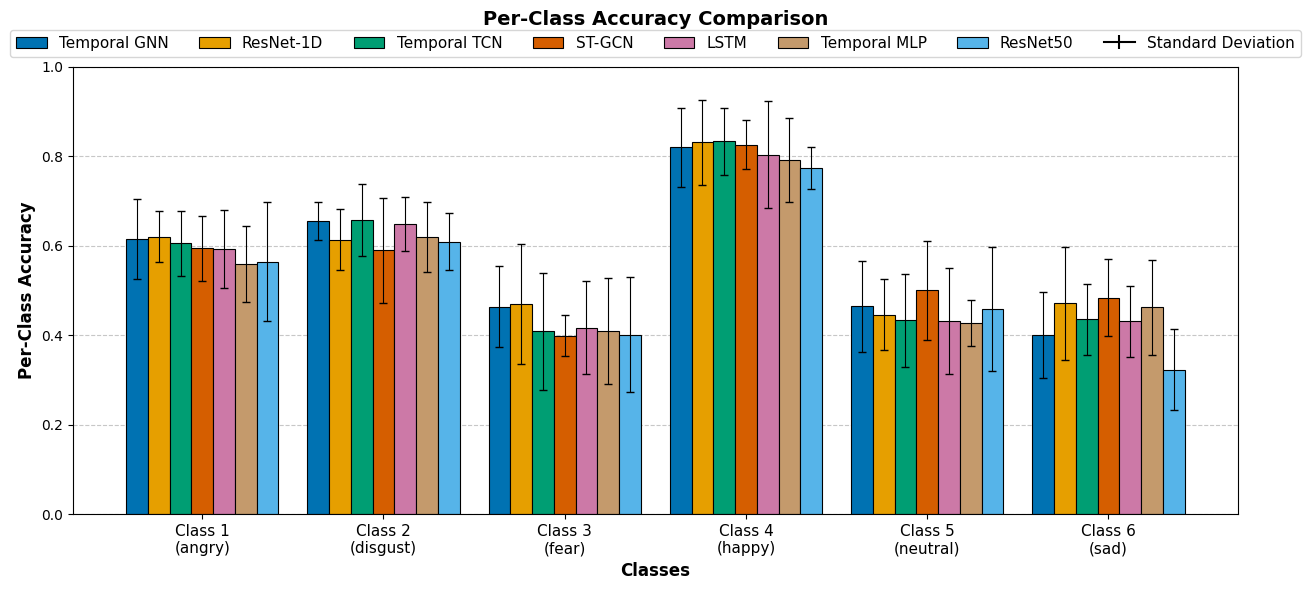}
\caption{Comparison of per-class accuracy for each architecture.}
\label{fig:per_class}
\end{figure*}

The per-class analysis (Fig.~\ref{fig:per_class}) provides a detailed breakdown of the models' predictive capabilities across different emotional categories, including the variance (standard deviation) across the 5 cross-validation folds. The data reveal that the overall underperformance of the raw-image baseline (ResNet50) is consistent across nearly all emotions. Notably, in the ``happy'' class---which yielded the highest accuracies for all architectures (around 80\%)---ResNet50 still recorded the lowest performance in the group. The disparity becomes even more pronounced in the most challenging classes, such as ``fear'' and ``sad''. For the ``sad'' class, ResNet50 achieved an accuracy of only about 32\%, significantly underperforming the landmark-based architectures, such as ST-GCN and ResNet-1D, which reached approximately 48\%. The standard deviations observed across all models reflect the variability inherent in the subject-independent cross-validation setup. In this scenario, the pixel-based baseline (ResNet50)---despite its larger parameter count and ImageNet pre-training---yielded lower accuracies across most emotional categories. Conversely, the architectures utilizing blendshapes and landmarks maintained higher average performances, indicating that these compact representations provided a more stable learning signal for this specific dataset.

\section{Ablation Studies}
\label{sec:ablationstudies}

As discussed previously (Section~\ref{sec:results}), the GNN architecture achieved the best balance between performance and parameter efficiency among the six evaluated approaches. Consequently, an ablation study was conducted to investigate the individual contribution of each input type (blendshape and landmark) to the model's final performance. To ensure a fair comparison, all ablation variants were trained and evaluated under the exact same 5-fold cross-validation setup and training conditions as the main experiments.

It is worth highlighting that because landmarks have an explicit spatial topology (MediaPipe's facial mesh tessellation), they can be modeled as a graph through the GINConv layers. Conversely, blendshapes have no spatial relationship among themselves, which makes it infeasible to construct an equivalent graph for this architecture. For this reason, the ablation variants required structural adaptations, while keeping the rest of the original modeling unchanged (hyperparameters, components, etc.):

\begin{itemize}
    \item \textbf{Blendshapes only:} The GINConv-based spatial module was entirely removed. Instead, the 52 blendshape coefficients are processed by the same dense branch (MLP 52 $\rightarrow$ 64 $\rightarrow$ 32, with ReLU activation) present in the original model. The results are fed directly into the Bidirectional GRU layer, without any interaction with geometric information.
    
    \item \textbf{Landmarks only:} The blendshape branch was removed. The original spatial flow was preserved---two GINConv layers (with batch normalization and ReLU activation) process the facial tessellation graph, followed by global mean pooling---and the resulting 64-dimensional embedding feeds directly into the Bidirectional GRU.
\end{itemize}

\begin{table}[htbp]
\caption{GNN performance by input modality. Bold values indicate the best results.}
\label{tab:ablation}
\begin{center}
\begin{tabular}{|l|c|c|}
\hline
\textbf{Variant} & \textbf{Accuracy (\%)} & \textbf{Macro F1-Score (\%)} \\
\hline
Complete model & $\mathbf{57.15 \pm 4.06}$ & $\mathbf{56.50 \pm 4.10}$ \\
Blendshapes only & $53.64 \pm {5.53}$ & $52.64 \pm 5.86$  \\
Landmarks only & $48.56 \pm {1.98}$ & $47.44 \pm 1.40$  \\
\hline
\end{tabular}
\end{center}
\end{table}

The results reveal an unequal contribution between the two modalities. Modeling with blendshapes alone resulted in an accuracy drop of approximately 3.5 percentage points relative to the full model, whereas modeling with landmarks alone caused a more severe drop of almost 9 percentage points. This result indicates that blendshapes carry most of the discriminative signal relevant to emotion classification, which is consistent with their nature: since they are coefficients that already semantically encode the activation of specific Action Units (ARKit standard), they provide the network with a pre-structured representation of the facial expression. Raw landmarks, in turn, require the GINConv module to learn, from scratch and from short temporal windows, which geometric relationships between points are relevant to each emotion---a more complex task when isolated.

In summary, the model using both input types outperforms both variants, indicating that, although blendshapes carry the primary discriminative signal, landmarks provide valuable complementary information, justifying the strategy of using both features across the six architectures evaluated in this study.

\section{Conclusion}
\label{sec:conclusion}

This work presented a comprehensive benchmark of six distinct deep learning architectures trained entirely from scratch for Facial Expression Recognition (FER), leveraging compact geometric representations (3D landmarks and ARKit blendshape coefficients) under privacy-preserving constraints via appearance abstraction and ephemeral video discarding. These models were rigorously compared against an ImageNet-pretrained ResNet50 convolutional baseline using a 5-fold subject-independent cross-validation protocol. By operating on ephemeral geometric primitives and discarding raw RGB video frames at capture, the proposed pipeline natively adheres to data minimization principles, significantly reducing data exposure risks under health privacy regulations.

The experimental results demonstrate that compact geometric representations achieve superior generalization on unseen individuals compared to dense pixel-based approaches. While the heavily parameterized ResNet50 baseline (25.9M parameters) experienced severe performance degradation under subject-independent validation ($51.97 \pm 6.00\%$), all six geometric models trained from scratch surpassed it. ResNet-1D achieved the highest overall accuracy ($58.06 \pm 6.59\%$), while the GNN attained competitive performance ($57.15 \pm 4.06\%$) with the lowest variance across folds and up to $124\times$ fewer parameters (208k). Furthermore, per-class evaluations demonstrated that geometric abstractions provide substantially more robust discriminative signals for subtle and complex expressions (such as sadness, where geometric models achieved $\sim 48\%$ accuracy compared to ResNet50's $32\%$).

Ablation studies established that while blendshapes encode the primary semantic information, facial landmarks provide complementary spatial topology that enhances overall classification performance. In practical clinical and assistive deployment, these geometric representations can be captured directly via native mobile frameworks (e.g., Apple ARKit) without persistent video storage, enabling real-time, low-power execution on edge devices. Future work will investigate attention-based transformer architectures and selective state-space models (Mamba) applied directly to continuous 3D facial point clouds.

\section*{Acknowledgments}
The authors would like to thank the support of the Research on Educational Adaptations for ASD, a project founded by PPI-Softex/MCTI by grant 01245.016126/2025-52 through the Brazilian Federal Government.

\bibliography{references.bib}
\bibliographystyle{IEEEtran}

\end{document}